\documentclass{article}
\usepackage{spconf,amsmath,graphicx}
\usepackage{multirow}
\usepackage{url}
\title{Video2IMU: Realistic IMU features and signals from videos}
\name{Arttu L\"{a}ms\"{a}$^{\star }$, Jaakko Tervonen$^{\star }$, Jussi Liikka$^{\star}$, Constantino Álvarez Casado$^{\dagger}$, Miguel Bordallo López$^{\star\dagger}$}

\address{$^{\star}$ VTT Technical Research Centre of Finland\\
        $^{\dagger}$ Center for Machine Vision and Signal Analysis (CMVS), University of Oulu, Finland }

\begin{document}
%\ninept
%
\maketitle

%
%%%%%%%%%%%  MAX: 150 WORDS--------------------------------
\begin{abstract}Human Activity Recognition (HAR) from wearable sensor data identifies movements or activities in unconstrained environments. HAR is a challenging problem as it presents great variability across subjects. Obtaining large amounts of labelled data is not straightforward, since wearable sensor signals are not easy to label upon simple human inspection. In our work, we propose the use of neural networks for the generation of realistic signals and features using human activity monocular videos. We show how these generated features and signals can be utilized, instead of their real counterparts, to train HAR models that can recognize activities using signals obtained with wearable sensors. To prove the validity of our methods, we perform experiments on an activity recognition dataset created for the improvement of industrial work safety. We show that our model is able to realistically generate virtual sensor signals and features usable to train a HAR classifier with comparable performance as the one trained using real sensor data. Our results enable the use of available, labelled video data for training HAR models to classify signals from wearable sensors.

\end{abstract}
%

%%---------------------------------------------------------------
%                         Introduction
%%---------------------------------------------------------------
\vspace{-2mm}
\section{Introduction}
\label{sec:intro}
\vspace{-2mm}
Human activity recognition (HAR) has been a popular field of interest among the research community. Due to its relatively accurate performance and high utility, HAR has found its way into several consumer products for noncritical use cases. These products, equipped with inertial motion unit (IMU) sensors, include mobile devices and wearable devices such as smart watches and wristbands. The devices use a combination of in-device processing and cloud services to produce information about the physical activity of the user, providing them with different context-adaptive services. 
Developing robust classifiers for detecting multiple activities is a challenging task that requires large amounts of labelled training data. The typical setup for obtaining this data is based on recording the IMU signals and labelling them in (semi) controlled sessions from multiple subjects, an approach that can be extremely laborious and presents serious limitations. On the other hand, video sharing platforms, such as YouTube, contain huge amounts of videos from various kinds of human physical activities, and they are usually easier to label based on their metadata or by simple visual inspection. Transforming this video information into a representation that could be directly used as a training material for IMU based classifers would enable access to enormous amounts of annotated training data, potentially improving the robustness of IMU-based HAR.
To directly use this video material for IMU-based activity classifiers, it must be transformed into IMU-like signals. In this paper, we show how extracting human movement from activity videos using pose estimation approaches produces signals that can be transformed into a representation that is directly usable by IMU-based classifiers. Our approach is based on a neural network model that uses a simple architecture to transform a sequence of 2D pose estimates from videos into realistic features equivalent to those obtained solely with IMU-signals. In addition, we show how the same neural network architecture could be used to extract the raw IMU-like signals themselves, by changing only the last layer. In this work, we concentrate solely on the acceleration signal, since it is the one predominantly used in HAR, but the method presented can be easily extended to other components of the IMU such as the magnetometer and gyroscope signals.

\vspace{-2mm}
\section{Related work}
\vspace{-2mm}
Although it is becoming relatively usual in visual computing \cite{augmentingVisualComputing} and computer vision tasks \cite{augmentingComputerVision} (including pose estimation \cite{augmentingPoseEstimation}),  using augmented, synthetic or generated data in IMU-based HAR has not yet been fully explored. A few small attempts at creating generated training data have been made. They include simulating different sensor positions and orientations using signal rotations \cite{augmentingSensorOrientation}, augmenting spectral data in the feature space \cite{augmentingHAR} or generating extra data by extrapolating time series \cite{augmentingTimeSeries}. All these approaches have presented mixed success and applicability, since the original labelled material used to derive the augmentations was still very limited \cite{limitedSensorData}.
In our work we aim at generating virtual sensor signals and features using only video sequences of human activities, a type of data that is abundant and easy to label. The closest work to ours is the one presented by Rey \textit{et al.} \cite{similarRey}. They experimented with the generation of sensor readings from monocular videos. Their approach is based on a regression model that predicts the total acceleration signal by using a sequence of pose estimates as an input of a typical residual convolutional network. The quality of their resulting signals is evaluated in a relatively simple activity recognition dataset using a classifier based on temporal convolutional blocks (TCN). Our approach aims instead at creating full training sets by generating sensor data directly from activity videos. We propose a U-net topology \cite{ronneberger2015u} to predict both individual feature values that could be directly used by an activity classifier, and 3-axis raw acceleration signals that can be directly compared to the original ones.

\vspace{-4mm}
\section{Experimental setup}
\vspace{-2mm}
To evaluate our approach, we performed our experimental analysis on the VTT-ConIoT dataset  \cite{Zenodo}\cite{VTT-ConIot}. This dataset recorded video data for 13 persons performing 15 different construction work-related activities for one minute. The subjects were wearing  three IMU-sensors at 100Hz. (located on the hip and near the left shoulder) 100 Hz sampling rate. The dataset provides synchronized IMU and video signals.

\begin{figure}[h]
\centering{\includegraphics[width=90mm]{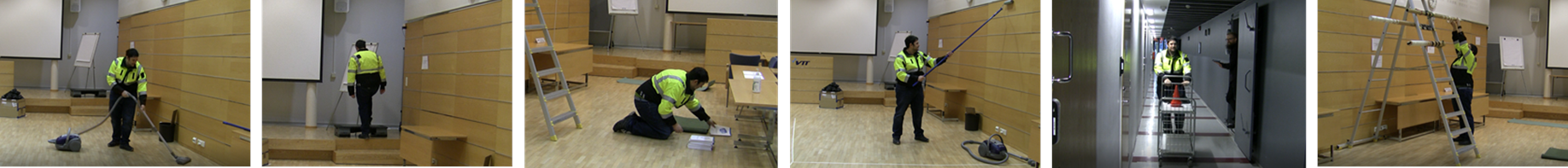}}
\vspace{-5mm}
\caption{Example of the 6-activity setup from the VTT-ConIot dataset. From left to right: Cleaning, Climbing, Floor Work, Painting, Walking and Hands Up}

\label{fig:activities}
\vspace{-5mm}
\end{figure}

In this work, we concentrate on the original dataset protocol described as the \textit{simple baseline}. This protocol defines six classes representing typical tasks in the construction setup: Cleaning, Climbing, Floor Work, Painting, Walking and working with Hands Up. Figure \ref{fig:activities} depicts and example frame of each of the tasks. In this setup, each 1-minute signal is segmented into sliding 2-second windows that are used as "activity samples" in both training and test sets. The performance is evaluated by constructing a model based on 7 different features (average, median, variance, mean, max, min, upper quartile and lower quartile) which are roughly based on the prior knowledge of their usability in the human activity. The features are calculated over 4 different signals obtained from the IMU placed on the hip (3-axis accelerometer, \textit{x,y,z} , and the total acceleration \textit{tot}), for a total of 28 features per sample. The cross-validation setup is Leave-one-subject-out (LOSO).
%COMMENT: The main point here is to include this simple baseline in the paper describing the dataset so we do not need to justify it here too much, since there is not a lot of space.

\section{Acceleration extraction from monocular videos}
\vspace{-2mm}
Processing videos to obtain a sequence of poses could provide some of the required positional information to estimate the acceleration of a human articulation or joint. Available open source pose estimation models include OpenPose \cite{openpose}, DeepPose \cite{deeppose} and Detectron2 \cite{detectron2}. Human pose information from monocular videos is essentially two-dimensional and expressed in absolute terms depending on the camera position, acquired at rates from 15 to 60 fps. On the other hand, IMU-based acceleration signals are three-dimensional (\textit{x,y,z}) and related to the position of the sensor itself, and they are recorded at much higher rates (50-150 Hz.). This calls for a preprocessing stage that allows us to match both signals.

The video preprocessing starts by extracting the human poses from each individual frame using a state-of-the-art human pose detector (Detectron2 \cite{detectron2}), which provides us with poses composed of 17 keypoints that represent articulations in the human body, in a usual format \cite{COCO}. 

We estimate the body center by interpolating between two keypoints located on both sides of the hip, the closest point to the IMU-sensor location (lower back center). This body position was selected as a reference point used in the experiments. To obtain a third spatial dimension of the point, we estimate the Z-coordinate using the size of the projected pose, since poses closer to the camera will look larger and poses far from the camera will look smaller. This distance was estimated by using the maximum and minimum size values of the dataset for each individual while standing (respectively, closer and farther distances to the camera), which allowed us to estimate the height of the subject and normalize the size according to it.

These coordinates represent the spatial displacements of the subjects while performing activities. We computed 3-axis accelerations of the reference point by calculating the second derivative of the coordinates across frames. This resulted in acceleration signals in the video coordinate space. This signal was then further processed with a low-pass filtering with a cut-off frequency of 12 Hz, to eliminate the high-frequency noise introduced by the pose estimation jittering and human detection errors.

Similarly, to match the characteristics of the acceleration signal obtained from the videos, the acceleration signals obtained from the real IMU-sensors were downsampled to 25 Hz (to match the video frame rate) and then low-pass filtered at 12Hz. This bandwidth reduction has shown to not affect the HAR classification process, since human activities do not present frequencies over 20 Hz, while most discriminative components are well below 6Hz \cite{frequencyHAR}\cite{frequency2HAR}.

\section{Transformation model}
\vspace{-2mm}

\begin{figure}[ht]
\centering{\includegraphics[width=60mm]{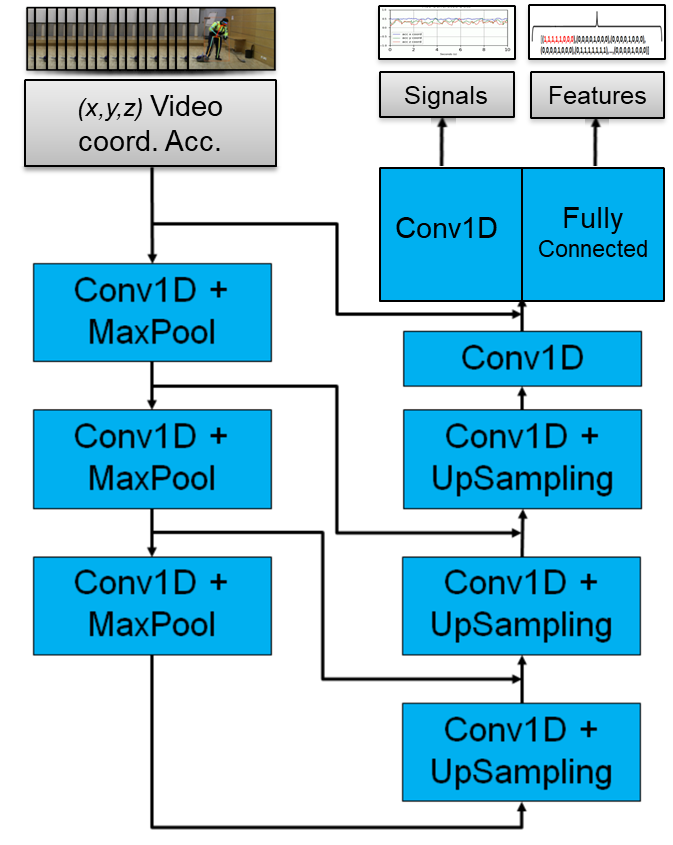}}

\vspace{-2mm}
\caption{Structure of the model used in the transformation. Output layers: 1-D convolutional filter for raw signals, fully connected layer for ndividual features.}

\label{fig:model}
\vspace{-2mm}
\end{figure}

The model used in the transformation between the pose estimation and IMU-like signals and features is based on the U-net topology \cite{ronneberger2015u} and consists of an encoder, decoder and skip-connections between layers that have feature maps of similar size. U-net has been previously used successfully in e.g., image segmentation tasks. The structure of the model used in the experiments is presented in Figure \ref{fig:model}. First, three convolutional blocks of the model (encoder) perform down-sampling on the data (on the left hand side of figure) while the next three blocks perform up-sampling (decoder). The down- and up-sampling blocks are connected with lateral skip connections that provide additional information for the convolutional/upsampling blocks. All our transformation models use the same architecture, with the exception of the output layer. This layer of the model is set as a 1-dimensional convolution layer for the generation of raw signals and as a fully connected layer for the generation of individual features. The selected optimizer is based on Adadelta's default parameters in \textit{keras.io}, using mean-absolute-error as the loss function. Each model was trained for 250 epochs with an early stopping configuration.

\section{Generation of virtual sensor signals}
\vspace{-2mm}
As a first step to use the generated data into HAR classifiers, we experiment with the generation of raw signals from the components obtained from video data. In this context, we generate a set of 2-second segments with 1-second overlap of the raw signals corresponding to (\textit{x,y,z}) axes and the total acceleration in an independent manner, constructing one transformation model per signal that follows the description detailed in the previous section.

To evaluate the correctness of these generated signals, we simply compare each segment belonging to the real and the generated ones, by computing the errors measured sample by sample along the segment. The errors are then expressed in terms of Mean Squared Error (MSE). The errors of the signal generation in comparison with the real IMU-based ones are depicted in Table \ref{tab:signals}.
\begin{table}[h]
\small
\centering
\vspace{-1mm}
\caption{Average MSE for each window and activity }
\label{tab:signals}
\begin{tabular}{|l|c|c|c|c|c|c|c|}
\hline
Signal    & \tiny Cleaning & \tiny Climbing & \tiny FloorWork & \tiny Painting & \tiny Walking &  \tiny HandsUp & Mean \\ \hline
$x_{(t)}$   &      0.16 &                     0.28 &        0.20 &      0.09 &   0.27 &      0.09 & 0.18 \\\hline
$y_{(t)}$   &      0.16 &                     0.16 &        0.20 &      0.11 &   0.19 &      0.08 & 0.15 \\\hline
$z_{(t)}$   &      0.57 &                     0.44 &        0.95 &      0.49 &   0.48 &      0.47 & 0.56 \\\hline
$tot_{(t)}$ &      0.13 &                     0.26 &        0.07 &      0.07 &   0.29 &      0.07 & 0.15 \\\hline
\end{tabular}
\end{table}

The results show that we are able to generate reasonably accurate signals for each axis and for the total acceleration without clear differences among different activities. However, we can note that z-axis present a significantly larger error than the other signals. We hypothesize that this is due to the changes in the gravity direction, which for the typical sensor setup and a standing person, corresponds exactly to that axis. This can be defended by, for example observing that the activity \textit{FloorWork} shows the larger error, and it is the activity where the gravity direction has more distinct and constant change, since the subjects are crouched in an inclined position and do not stand up during the whole activity.

\section{Generation of virtual sensor features}
\vspace{-2mm}
The next step was to create models that are able to predict specific statistical features calculated from the IMU signal by using information obtained only from video data. These statistical features can then be directly used for HAR classification, without the need of creating accurate signals.

In this context, analogously to signal generation, we create a set of models whose output is a single value that corresponds to each IMU-feature as it is used by the classifiers. We output this value by changing in our architecture the last 1x1 convolutional layer to a fully connected layer that provides a single value per sample. The set of features is selected to correspond to the simple baseline of the VTT-ConIot dataset classification task~\cite{VTT-ConIot}. The set of models estimates seven statistical features: average \textit{(avg)}, median \textit{(med)}, variance \textit{(var)}, 25th percentile \textit{(lq)}, 75th percentile \textit{(uq)}, minimum \textit{(min)} and maximum \textit{(max)}. Each feature is computed in four different axes (\textit{x, y, z} and total acceleration \textit{(tot)} computed as the L2 norm of the axes). Each feature is estimated using 2-second segments, and the features were generated using leave-one-subject-out (LOSO) validation.

This process produced a vector with 28 feature values for each two second time widow. The mean-squared-errors of each generated feature in comparison to the real IMU feature values are shown in Table \ref{tab:mses}. \hfill\break

% Please add the following required packages to your document preamble:
% \usepackage{graphicx}

\begin{table}[h]
\small
\centering
\vspace{-2mm}
\caption{Average MSE for each feature and activity}
%\begin{tabular}{|@{\hskip2pt}l@{\hskip2pt}|@{\hskip2pt}c@{\hskip2pt}|@{\hskip2pt}c@{\hskip2pt}|@{\hskip2pt}c@{\hskip2pt}|@{\hskip2pt}c@{\hskip2pt}|@{\hskip2pt}c@{\hskip2pt}|@{\hskip2pt}c@{\hskip2pt}|@{\hskip2pt}c@{\hskip2pt}|}
\begin{tabular}{|@{\hskip3pt}l@{\hskip3pt}|@{\hskip3pt}c|@{\hskip3pt}c|@{\hskip3pt}c|@{\hskip3pt}c|@{\hskip3pt}c|@{\hskip3pt}c|@{\hskip3pt}c|}
\hline
Feat    & \tiny Cleaning & \tiny Climbing & \tiny FloorWork & \tiny Painting & \tiny Walking &  \tiny HandsUp & Mean \\ \hline
$x_{avg}$   & 0.023      & 0.007      & 0.155      & 0.008      & 0.016     & 0.021     & 0.038  \\ \hline
$x_{med}$   & 0.026      & 0.008      & 0.144      & 0.008      & 0.016     & 0.024     & 0.038  \\ \hline
$x_{var}$   & 0.049      & 0.092      & 0.051       & 0.024     & 0.043     & 0.033     & 0.049  \\ \hline
$x_{lq}$    & 0.059      & 0.041      & 0.141      & 0.018      & 0.041     & 0.042     & 0.057  \\ \hline
$x_{uq}$   & 0.025      & 0.025      & 0.157      & 0.010      & 0.047     & 0.029     & 0.049  \\ \hline
$x_{min}$   & 0.226      & 0.607     & 0.261      & 0.105       & 0.197    & 0.137    & 0.256 \\ \hline
$x_{max}$   & 0.285      & 0.670     & 0.289      & 0.129       & 0.437    & 0.196    & 0.334 \\ \hline
$y_{avg}$   & 0.037      & 0.029      & 0.185      & 0.038      & 0.044     & 0.048     & 0.064  \\ \hline
$y_{med}$   & 0.043      & 0.027      & 0.183      & 0.041      & 0.037     & 0.047     & 0.063  \\ \hline
$y_{var}$   & 0.006      & 0.007      & 0.002       & 0.002     & 0.007     & 0.002     & 0.004  \\ \hline
$y_{lq}$    & 0.056      & 0.036      & 0.220      & 0.045      & 0.043     & 0.058     & 0.076  \\ \hline
$y_{uq}$    & 0.044      & 0.039      & 0.143      & 0.044      & 0.045     & 0.042     & 0.060  \\ \hline
$y_{min}$   & 0.200      & 0.378     & 0.345      & 0.090       & 0.331    & 0.145    & 0.248 \\ \hline
$y_{max}$   & 0.192      & 0.427     & 0.123      & 0.091       & 0.204    & 0.082     & 0.187 \\ \hline
$z_{avg}$   & 0.313      & 0.178     & 0.850      & 0.269       & 0.173    & 0.301    & 0.347 \\ \hline
$z_{med}$   & 0.307      & 0.168     & 0.829      & 0.298       & 0.127    & 0.286    & 0.336 \\ \hline
$z_{var}$   & 0.009      & 0.010      & 0.004       & 0.002    & 0.013     & 0.004     & 0.007  \\ \hline
$z_{lq}$    & 0.332      & 0.188     & 0.979      & 0.275       & 0.236    & 0.293    & 0.384 \\ \hline
$z_{uq}$    & 0.287      & 0.162     & 0.801      & 0.258       & 0.125    & 0.311    & 0.324 \\ \hline
$z_{min}$   & 0.428      & 0.586     & 1.229     & 0.313        & 0.614    & 0.361    & 0.589 \\ \hline
$z_{max}$   & 0.376      & 0.458     & 0.839      & 0.278       & 0.361    & 0.464    & 0.463 \\ \hline
$tot_{avg}$ & 0.008      & 0.008      & 0.004       & 0.003      & 0.027     & 0.007     & 0.010  \\ \hline
$tot_{med}$ & 0.006      & 0.007      & 0.004       & 0.003      & 0.022     & 0.008     & 0.008  \\ \hline
$tot_{var}$ & 0.028      & 0.074      & 0.014       & 0.011      & 0.045     & 0.018     & 0.032  \\ \hline
$tot_{lq}$  & 0.019      & 0.029      & 0.007       & 0.008      & 0.043     & 0.015     & 0.020  \\ \hline
$tot_{uq}$  & 0.025      & 0.030      & 0.010       & 0.012      & 0.075     & 0.015     & 0.028  \\\hline
$tot_{min}$ & 0.104      & 0.276     & 0.055       & 0.041      & 0.152    & 0.062     & 0.115 \\ \hline
$tot_{max}$ & 0.386      & 0.910     & 0.168      & 0.137     & 0.547    & 0.190    & 0.390  \\ \hline
Mean        & 0.139      & 0.196     & 0.293      & 0.091      & 0.145    & 0.116    & 0.163 \\ \hline

\end{tabular}\vspace{-3mm}
\label{tab:mses}
\end{table}

The results show again that we are able to generate individual features with reasonable accuracy across different activities, without clear differences. Again, it can be observed that the generation of features in the z-axis is more challenging, especially for features that measure average or median values, while the variance error is similar to the one obtained for the other axes. We argue that our model is able to infer the energy and variations of the individual signals while obtaining its average values (that depend on the sensor orientation) is more difficult. These are to be expected since the input values obtained from the videos and poses do not provide any information on the gravity components or directions.  

\vspace{-2mm}
\section{Classification of human activities}
\vspace{-2mm}
Finally, we evaluate the usability of the models and their generated features by using classifiers. For a fair comparison, we train three different classifiers using three different types of input data. The first classifier (IMU input) is trained using only real IMU data obtained from real sensors. According to the protocol described before, 28 features were calculated and standardized (by subtracting the mean and dividing with standard deviation) from the real IMU signals. These features are used to train a random forest classifier to classify six activities. A second classifier (Generated input) is trained by using the features generated from our previously described generation model that produced already standardized features. This classifier is trained on videos that do not include the subject in the testing. Finally, for comparative purposes, a third naive classifier (Video input) is trained by using the same type of standardized features that are directly computed from the accelerations corresponding to the poses in the videos, without an intermediate generation model.  

We run inference on all three classifiers in the same way. For testing, we used only real IMU data from unseen subjects, and we strictly followed a leave-one-subject-out protocol where none of the trained models (both for generation and classification) had access to the test subject.The classifiers, trained with different data types (real, generated, and naive-video) are evaluated in terms of mean classification accuracy. Results for all three classifiers are shown in Table \ref{tab:loo}.

\begin{table}[ht]
\small
\centering
\vspace{-4mm}
\caption{LOSO validation results of different IMU classifiers.}
\begin{tabular}{|c|c|c|c|}
\hline
        & Accuracy              & Accuracy                    & Accuracy     \\ 
Subject & IMU input            & Generated input            & Video input               \\ \hline
S1      & 0.495                 & 0.489                       & 0.290                      \\ \hline
S2      & 0.461                 & 0.395                       & 0.207                      \\ \hline
S3      & 0.415                 & 0.375                       & 0.254                      \\ \hline
S4      & 0.466                 & 0.379                       & 0.328                      \\ \hline
S5      & 0.493                 & 0.518                       & 0.217                      \\ \hline
S6      & 0.518                 & 0.453                       & 0.341                      \\ \hline
S7      & 0.623                 & 0.622                       & 0.209                      \\ \hline
S8      & 0.567                 & 0.525                       & 0.262                      \\ \hline
S9      & 0.392                 & 0.304                       & 0.206                      \\ \hline
S10     & 0.589                 & 0.547                       & 0.231                      \\ \hline
S11     & 0.560                 & 0.547                       & 0.244                      \\ \hline
S12     & 0.582                 & 0.554                       & 0.302                      \\ \hline
S13     & 0.621                 & 0.547                       & 0.303                      \\ \hline
\textbf{Mean} & \textbf{0.521}  & \textbf{0.481}              & \textbf{0.261}             \\ \hline
\end{tabular}
\label{tab:loo}
\end{table}

The results clearly show the usefulness of the intermediate generation model. The classifier trained on only real IMU data, following the simple baseline of the dataset, obtains an accuracy of about 52\%, while the naive classifier trained directly from video pose coordinates and tested on real IMU-data achieves about half (26\%). However, the model trained with the intermediate feature generation model achieves a 48\% accuracy, a performance comparable to the classifier trained on real IMU data. We believe that these results show that our Video2IMU generation models are able to add IMU-like characteristics to the acceleration signal extracted from pose estimation coordinates, in a way that can generalize to unseen individuals.

\vspace{-2mm}
\section{Discussion}
\vspace{-2mm}
We have shown that a relatively simple transformation model to generate IMU-like signals and features from activity videos can be directly utilized to train HAR classifiers that can perform inference on real sensor data with reasonable accuracy. 
Our selected setup uses a simple baseline where the short window size, reduced number of features, and a single sensor location (placed in the hip) presents a classification accuracy that is relatively low, even for the IMU-based classifier that uses real sensor data. We believe that this is not the optimal setup for recognizing certain activities where movement mostly occurs on the hands. Future work will focus on studying alternative sensor locations and the limits for the generalization of our generation method to unseen activities.

% References should be produced using the bibtex program from suitable
% BiBTeX files (here: strings, refs, manuals). The IEEEbib.bst bibliography
% style file from IEEE produces unsorted bibliography list.
% -------------------------------------------------------------------------
\bibliographystyle{IEEEbib}
\bibliography{strings,refs}

\end{document}